\documentclass[manuscript,screen]{acmart}
\usepackage{color}
\usepackage{multirow}
\usepackage{mathtools}
\usepackage{amsfonts}
\usepackage{algorithmic}
\usepackage{kotex}
\definecolor{pp}{rgb}{0.6,0.0,0.6}

\AtBeginDocument{%
  \providecommand\BibTeX{{%
    \normalfont B\kern-0.5em{\scshape i\kern-0.25em b}\kern-0.8em\TeX}}}

\setcopyright{acmcopyright}
\copyrightyear{2018}
\acmYear{2018}
\acmDOI{XXXXXXX.XXXXXXX}

\acmConference[RecSys '23]{Make sure to enter the correct
  conference title from your rights confirmation email}{June 03--05,
  2018}{Woodstock, NY}
\acmBooktitle{RecSys '23: ACM Symposium on Neural Gaze Detection,
 June 03--05, 2018, Woodstock, NY} 
\acmPrice{15.00}
\acmISBN{978-1-4503-XXXX-X/18/06}

\begin{document}

\title{Lightweight Boosting Models for User Response Prediction Using Adversarial Validation}

\author{Hyeonwoo Kim}
\authornote{Both authors contributed equally to this research.}
\email{choco_9966@upstage.ai}
\affiliation{%
  \institution{Upstage AI Research}
  % \streetaddress{P.O. Box 1212}
  \country{Republic of Korea}
}

\author{Wonsung Lee}
\authornotemark[1]
\email{wonsung.lee@upstage.ai}
\affiliation{%
  \institution{Upstage AI Research}
  % \streetaddress{P.O. Box 1212}
  \country{Republic of Korea}
}


\begin{abstract}
The ACM RecSys Challenge 2023, organized by ShareChat, aims to predict the probability of the app being installed. This paper describes the lightweight solution to this challenge. We formulate the task as a user response prediction task. For rapid prototyping for the task, we propose a lightweight solution including the following steps: 1) using adversarial validation, we effectively eliminate uninformative features from a dataset; 2) to address noisy continuous features and categorical features with a large number of unique values, we employ feature engineering techniques.; 3) we leverage Gradient Boosted Decision Trees (GBDT) for their exceptional performance and scalability. The experiments show that a single LightGBM model, without additional ensembling, performs quite well. Our team achieved ninth place in the challenge with the final leaderboard score of 6.059065. Code for our approach can be found here: \href{https://github.com/choco9966/recsys-challenge-2023}{https://github.com/choco9966/recsys-challenge-2023}.
\end{abstract}

\begin{CCSXML}
<ccs2012>
   <concept>
       <concept_id>10002951.10003317.10003347.10003350</concept_id>
       <concept_desc>Information systems~Recommender systems</concept_desc>
       <concept_significance>500</concept_significance>
       </concept>
 </ccs2012>
\end{CCSXML}

\ccsdesc[500]{Information systems~Recommender systems}

\keywords{ACM RecSys Challenge 2023, User Response Prediction, Adversarial Validation, Gradient Boosting Decision Trees, CTR Prediction}


\maketitle

\section{Introduction}
% CTR 예측 task 개요. model-centric 관점. from shallow to deep
% 생략 , in terms of a model-centric perspective
User response prediction is challenging due to the inherent nature of the task, such as data sparsity, noisy and incomplete data, concept drift, and temporal dynamics. Researchers have explored diverse approaches to address this issue and improve prediction accuracy, including logistic regression (LR) \cite{google-ctr}, decision trees \cite{fb-ctr}, factorization machines (FM) \cite{ffm}, and deep learning-based methods \cite{dcn, din, dmt}. Deep learning models for user response prediction have recently gained popularity due to their ability to provide an inductive bias suitable for the input data type and capture intricate non-linear relationships in complex high-dimensional data spaces. However, despite the success of deep learning in predicting user responses, it is also widely known that tree-based methods often outperform neural networks in tabular data prediction tasks, especially in competitions.

The goal of ACM RecSys Challenge 2023 is to estimate the probability of the app being installed. This task can be formulated naturally as a click-through rate (CTR) or conversion rate (CVR) prediction problem, which is a type of user response prediction. For rapid prototyping for the task, we propose a lightweight solution including the following steps: 1) given a dataset consisting of user and ad features, we employ adversarial validation to exclude non-informative features effectively; 2) we perform feature engineering to deal with noisy continuous features and categorical features with high cardinality; 3) we utilize GBDT due to their superior performance and scalability. Our contributions are as follows. First, we provide a lightweight solution that combines adversarial validation and a set of feature engineering techniques. Second, we show that even a single LightGBM model without an additional ensembling performs quite well in user response prediction, ranking ninth in the leaderboard.

%The rest of the paper is organized as follows. We first introduce related work including CTR prediction and competitive data science. Then we 

\section{Related Work}
\subsection{User Response Prediction}
% 초기 연구들, Google's LR with FTRL-Proximal, FB's GBDT+LR, Criteo's FFM
Early studies that tackled user response prediction tasks include the following significant studies (mainly CTR prediction): \cite{google-ctr, fb-ctr, ffm}. Google \cite{google-ctr} presented a regularized LR model based on an FTRL-Proximal algorithm for massive-scale sponsored search advertising. They explored several memory-saving guidances essential for building an industrial-scale CTR prediction system, such as bloom filter, fewer bit encoding, and train data sub-sampling. Another work \cite{fb-ctr} published by Facebook employed a combination of decision trees and LR. The boosted decision tree performs non-linear supervised feature encoding and feeds the encoded features as an input to LR. They also stressed the importance of train data sub-sampling, negative down-sampling, and model calibration to handle large-scale real-world data. Field-aware FM \cite{ffm}, which is a variant of FM \cite{fm}, also showed an impressive performance, showing the critical importance of capturing predictive feature interactions.

% 최근의 딥러닝 for CTR pred. 연구들. Google's DCN, Alibaba's DIN, JD.com's DMT
Recent years have seen a growth of interest in the adoption of deep learning-based models for CTR prediction. Due to the superior ability of representation learning with multiple levels of abstraction, the use of deep neural networks (DNN) in industrial-scale CTR prediction is gradually becoming the industry standard \cite{dcn, din, dmt}. The Deep \& Cross Network (DCN) \cite{dcn} adopts a memory-efficient cross network that explicitly learns predictive cross features without manual feature engineering. This simple yet effective architecture with significantly fewer parameters outperforms other baselines, including DNN, FM, and LR. Alibaba proposed a Deep Interest Network (DIN) \cite{din}, which utilizes a local activation unit, similar to an attention mechanism, to dynamically learn the representation of user interests based on their past behaviors related to a specific advertisement. Deep Multifaceted Transformers (DMT) \cite{dmt}, which models multiple types of behaviors (e.g., click, cart, and order) with multiple Transformers \cite{trm}, has been successfully deployed to JD.com, one of the largest E-commerce sites in the world. DMT also shows that properly adopting multi-task learning that recognizes the user's different objectives can significantly improve real-world online performance metrics such as CTR, Conversion Rate (CVR), and Gross Merchandise Volume (GMV).
% 원래 학술 논문에서는 이 파트에 우리 모델과의 차이점을 언급하지만, 우리는 특별히 CTR prediction 연구 관점에서는 특별히 언급할만한 내용이 없으므로 생략함

%Meanwhile, in terms of a data-centric perspective, various feature engineering methods also have been proposed to improve the performance of CTR prediction. \ws{Google, FB, Alibaba 등에서 어떤 feature eng. 기법 사용했는지 정리 + 최근의 AutoML 관점 정리하기, user historical feats이 중요하더라... Youtube Ranking에서 어떤 feature가 중요했는지... AutoML 측면에서 AutoFIS, AutoCTR? 얘기}

\subsection{Competitive Data Science}
% kaggle, recsys challenge 와 같은 데싸 경쟁 대회 이야기 하면서 우리가 tree-based model + feat eng.에 집중한 이유를 뒷받침 하기 위한 문단
Data science competitions have evolved from a niche community of passionate participants to a widely popular platform attracting millions of data scientists worldwide \cite{kaggle}. In 1997, the KDD Cup, one of the earliest data science competitions, was held. In 2006, the Netflix Prize \cite{netflix} was held, significantly impacting the development of personalized recommender systems and collaborative filtering. Since then, various data science competitions focusing on personalized recommendations and user response prediction have been held at academic conferences and competitive data science platforms.

We observed a wide variety of methodologies being used for personalized predictions in competitions, from TF-IDF to advanced deep learning architectures. Surprisingly, several top-ranking solutions utilized tree-based models. For example, in 2017, 2018, 2019, 2020, and 2022 ACM RecSys challenges\footnote{\href{https://recsys.acm.org/challenges}{https://recsys.acm.org/challenges}}, the winning solutions utilized GBDT such as LightGBM \cite{lightgbm}, CatBoost \cite{catboost}, and XGBoost \cite{xgboost} with substantial feature engineering. Also, the winning solution\footnote{\href{https://www.kaggle.com/competitions/h-and-m-personalized-fashion-recommendations/discussion/324070}{https://www.kaggle.com/competitions/h-and-m-personalized-fashion-recommendations/discussion/324070}} of the H\&M Personalized Fashion Recommendations competition, recently held at Kaggle, took advantage of different kinds of decision tree models (LightGBM and CatBoost) as a ranker module. Although deep learning has achieved tremendous success in the computer vision and natural language processing domains, it is known that tree-based methods often outperform neural networks in tabular data prediction tasks, especially when dealing with skewed distributions, heavy-tailed feature distributions, and dataset irregularities \cite{tree-vs-dl1, tree-vs-dl2, tree-vs-dl3}.

\section{Problem Formulation}
\subsection{Dataset Description}
The data available for the challenge was provided by ShareChat\footnote{\href{https://sharechat.com/recsys2023}{https://sharechat.com/recsys2023}}, India's largest homegrown social media company with over 400M MAUs across all its platforms. The dataset consists of 10M users who visited the ShareChat and Moj apps over three months. The organizers preprocessed the dataset to have ten impressions for each user. The objective of the challenge is to predict the probability of the app being installed. Each row of the dataset consists of user and ad features. The train data consists of subsampled history from the past $22$ days, and the target variable is the probability of the app being installed on the $23$rd day.

%The user features include demographic features, content preference embeddings, and app affinity embeddings. The ad features include ad categorical features and the corresponding ad embedding. In addition, to capture the historical interaction between users and ads, count features that represented the user interaction with ads, advertisers, and categories of advertisers over different lengths of a time window, are provided.

\subsection{Evaluation}
The objective of the challenge is to predict whether the user will install or not for a given ad impression. The more accurately the probability is estimated, the higher the expected revenue of the platform. The metric used in the challenge was Normalized Cross Entropy (NCE):
\begin{equation} \begin{aligned}
NCE = \frac{-\frac{1}{N} \sum_{i=1}^{N} (\frac{1+y_i}{2} \log{p_i} + \frac{1-y_i}{2} \log{(1-p_i)}) }{-(p \log{p} + (1-p)  \log{(1-p)})},
\end{aligned} \end{equation}
where $N$, $p_i$, $p$, and $y_i \in \{ -1, 1 \}$ are the number of the dataset, the estimated probability of the app being installed, the average empirical probability of installation, and the label, respectively.

NCE is calculated by dividing the average log loss per impression by the average log loss per impression that would occur if the model predicted the background CTR (in our case, CVR) for every impression. In other words, NCE is the predictive log loss normalized by the entropy of the empirical probability of installation. The lower the NCE value, the better the model performance. In our experiments, we use log loss as a proxy evaluation metric for local validation because we can not access the ground-truth empirical CVR, $p$.

%NCE is also used to calculate Relative Information Gain (RIG): $RIG = 1- NCE$, which might be more descriptive and convenient for presentation \cite{fb-ctr}.

%$p_i$ is the estimated probability of install for a particular ad, $p$ is the average empirical probability of install and $y_i \in \{ -1, 1 \}$ is the label.

%, $N$, $p_i$ are the number of the dataset, the estimated probability of install
% ref: http://chuckcode.com/post/fb_practical_lessons/

\section{Data Preparation \& Preprocessing}

\begin{figure}[t!]
\centering
\includegraphics[width=0.6\linewidth]{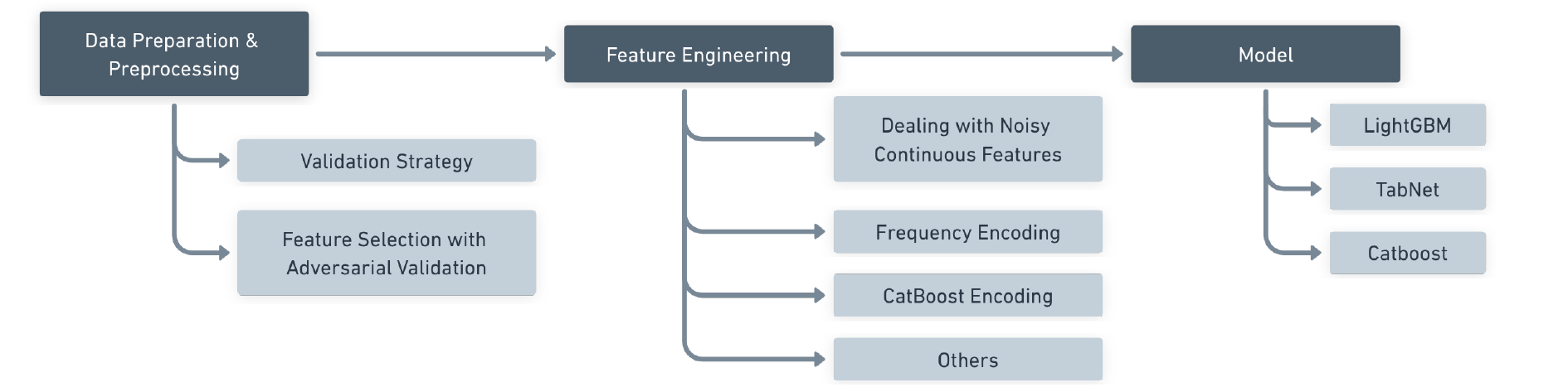}
\caption{Flow chart of our lightweight pipeline}
\label{fig:diagram}
\end{figure}

\subsection{Local Validation Strategy}
The dataset includes temporal information as the $f_1$ variable, which indicates the date for each row. We aim to predict $p_i$ as accurately as possible for test data with $f_1=67$. Therefore, to reflect recent temporal trends and to make the validation set's distribution mimic the test set's distribution, we select the data with $f_1=66$ as a local validation set. The entire procedure of the pipeline is shown in Fig. \ref{fig:diagram}.

\subsection{Adversarial Validation}
\label{sec:cov-shift}

\begin{figure}[t!]
\centering
\includegraphics[width=0.9\linewidth]{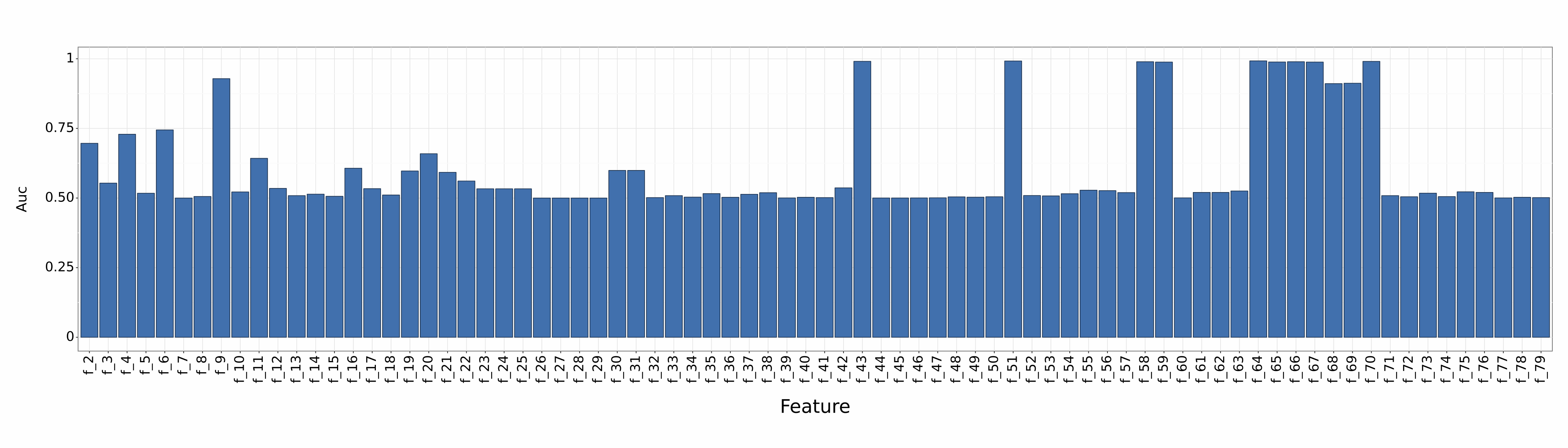}
\caption{AUC metrics of each feature in adversarial validation for detecting a covariate shift.}
\label{fig:adv-validation}
\end{figure}

% Concept drift: Concept drift refers to the situation where the statistical properties of the target variable or the relationships between features and the target variable change over time.
% Covariate shift: Covariate shift, also known as distribution shift, refers to a change in the distribution of the input features while keeping the target variable distribution constant.
% https://arxiv.org/pdf/2112.10078.pdf
In our preliminary experiments, we observed a severe discrepancy between the local validation score and the leaderboard score. In particular, this phenomenon was amplified when certain variables were added as features. This overfitting issue seems to be due to a covariate shift, the discrepancy between the distribution of train data and test data. To identify and address the problem, we employ an adversarial validation approach.

% \textbf{For example, in the case of the categorical variable $f_4$, the distribution of the train/valid period ($f_1 \leq 66$) and the test period ($f_1 = 67$) are significantly different: the cardinality value of the train/valid period and the test period are $633$ and $298$, respectively. Also, $293$ categories exist only in the train/valid period, and five categories exist only in the test period. Similarly, some continuous variables exhibit the covariate shift.

%Given a labeled training dataset $\mbD_{\mathtt{train}}=\{X_{\mathtt{train}}, y_{\mathtt{train}}\}$ and test dataset $\mbD_{\mathtt{test}}=\{ X_{\mathtt{test}} \}$
The adversarial validation approach can be used to detect and address the covariate shift. This approach involves training a binary classifier to determine whether a sample belongs to the training or test set. When the classifier's performance is close to random guessing ($\mathtt{AUC}=0.5$), it suggests that distinguishing between the training and test sets is challenging, indicating that the distribution of both is relatively consistent. On the other hand, if the classifier performs significantly better than random guessing, it suggests the discrepancy between the distribution of the training and test sets. We train an adversarial classifier for each feature to detect potential features that exhibit the covariate shift between train and test data. As shown in Fig. \ref{fig:adv-validation}, some variables have very high $\mathtt{AUC}$ scores, and these variables seem to show the covariate shift that may cause the gap between validation performance and test performance. Therefore, we can exclude these features from model training to reduce the inconsistency between the local validation and the leaderboard.

%\hw{Q. adv. validation으로 찾은 변수가 covariate shift를 일으켰는지, 아니면 concept drift와 같은 다른 종류의 data drift에 의한 것인지 구분할 수 있나요?}

%the discrepancy between the validation score and the leaderboard score. + overfitting with 

%f1을 기준으로 65까지를 train, 66을 valid로 두고 카테고리와 연속형 변수의 분포를 살펴보면 서로 다른 분포를 가짐 
%카테고리 f4 변수를 예시로 들면, train은 626, valid는 292의 cardinality를 가지고 이 중 314개의 값은 train에만 존재, 7개는 valid에만 존재함 
%마찬가지로 연속형 변수의 경우도 분포 자체가 다른 경우가 존재. 예를들어, f64 변수의 경우 box plot처럼 valid 자체가 전반적으로 높은 값을 가진 분포임을 확인할 수 있음 
%이러한 문제들은 Covariate shift 문제로 볼 수 있으며 학습된 모델이 미래에 들어올 데이터에 대해 정확한 예측을 못하게 할 수 있다. 
%우리는 좀 더 해당 문제를 분석하기 위해 Adversarial Validation 방법을 통해서 각 변수의 Covariate shift 정도를 파악했다. 

%Adversarial Validation은 주어진 데이터의 변수들을 가지고 모델이 Train과 Valid (혹은 Test)를 잘 구분하는지를 보는 방법이다. 
%실제, Train에는 target = 0 으로 두고 valid에는 target을 1로 두어서 랜덤하게 섞은 다음에, 모델이 주어진 피쳐를 통해 이 둘을 구분하는 능력이 있는지 보고 만약 구분을 잘 한다면 해당 피쳐는 train과 valid간의 차이가 있다고 해석할 수 있고, 그렇지 않다면 차이가 없다고 해석할 수 있다. 즉, 여기에서는 평가지표로 보는 auc가 0.5일 수록 해당 피쳐가 train과 valid가 같은 분포를 가진다는 것을 의미한다. 

%Fig \ref{fig:adv-validation} 결과를 분석해보면 일부 변수들의 경우 대부분의 변수가 0.5 근방의 값을 보이지만 일부 변수가 굉장히 높은 점수를 가지고 있고, 이런 변수들은 특히 학습 모델과 검증시점의 괴리를 일으킬 가능성이 높다고 판단내릴 수 있다. 

% 생략해도 될 듯?

\section{Feature Engineering}

Based on the analysis in \ref{sec:cov-shift}, we filter out the variables with $\mathtt{AUC} >= 0.75$ (indicating a potential covariate shift). Next, we encode categorical features to informative continuous values to deal with high cardinality and facilitate effective tree splits for GBDT models. In the following paragraphs, we introduce how to deal with noisy continuous features and how to encode categorical features. Please refer to our source code\footnote{\href{https://github.com/choco9966/recsys-challenge-2023}{https://github.com/choco9966/recsys-challenge-2023}} for other details of feature engineering.

\subsection{Dealing with Noisy Continuous Features}

\begin{figure}[t!]
\centering
\includegraphics[width=0.4\linewidth]{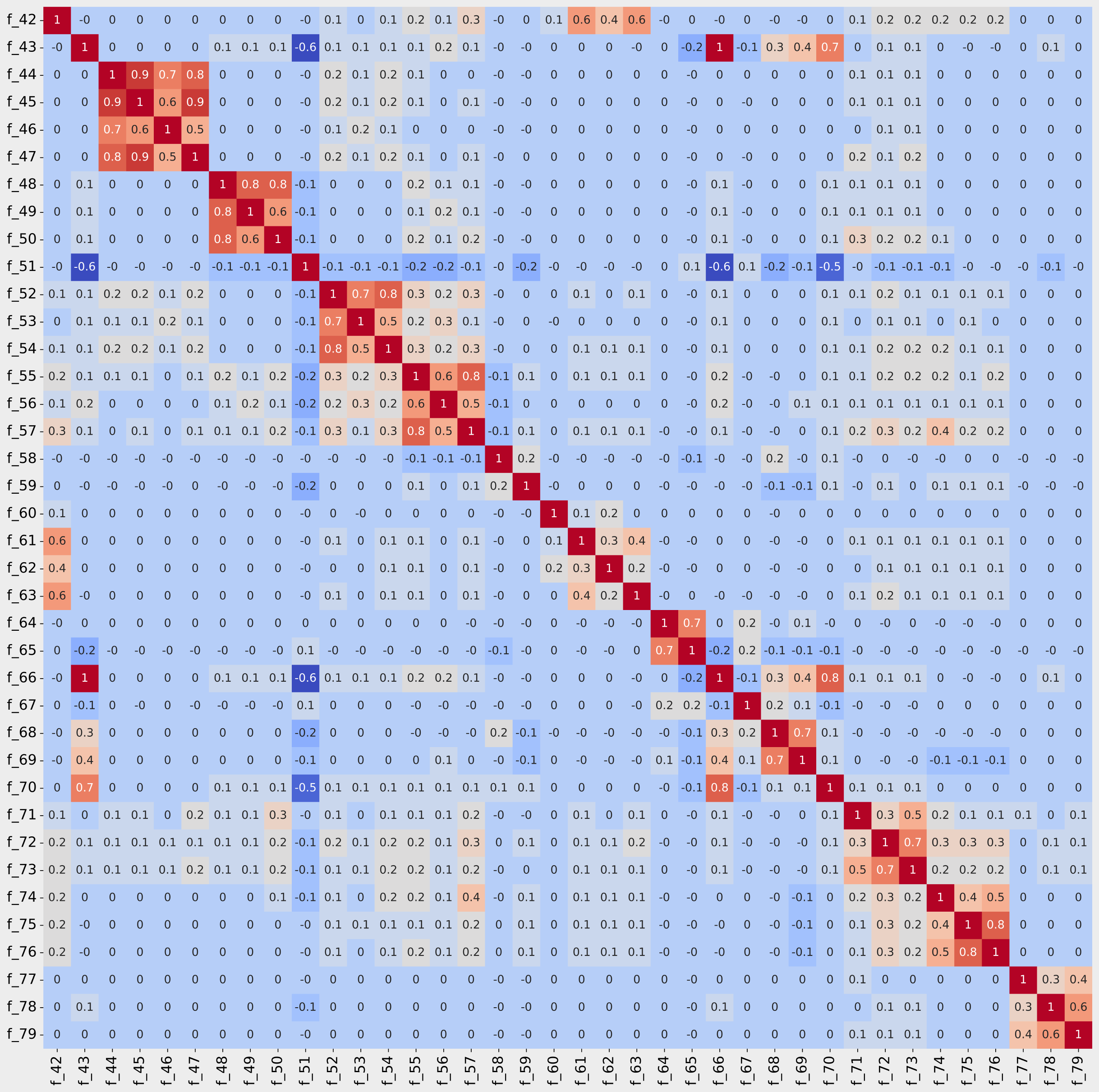}
\caption{Pairwise correlation plot for continuous features: $f_{42} \sim f_{79}$.}
\label{fig:correlation}
\end{figure}

% scaling? de-noising?...
% 이 부분 두 번째로 작은 값이 같다는게 어떤 의미일까요? organizer들이 임의로 더한 noise를 의미한다고 볼 수 있나요?
% 어떤 제목으로 만들면 좋으려나요. 
% noise injection이 있었으리라 추정함. PCA?
% count feature인데, discrete 하지 않고 continuous 특성을 보임.
% feature 별로 unique value들의 gap이 동일함 (e.g., 0, 0.0385640684536896, 0.0385640684536896*2, 0.0385640684536896*3, ...)
% 이걸로 나누줘서 count feature로 복원하는 효과 -> denoising + feature interaction 계산 시 더 정확함
Fig. \label{fig:corrleation} shows that some continuous variables exhibit a block structure indicating a strong correlation with each other. Based on this observation, we found a pattern presumed to be artificial noise injection. Specifically, some variables represent an arithmetic sequence of unique values. More concretely, given the ordered \textit{unique} values $\{v_1, v_2, ... v_M\}$ for a particular feature, where $v_{n-1} < v_{n}$ for $n=2, ..., M$, the general formula for the $n$-th term is as follows: $v_{n}=v_{1} + (n-1) \Delta$. For example, variable groups ($f_{42}$, $f_{52} \sim f_{57}$, $f_{74} \sim f_{76}$) and ($f_{44} \sim f_{50}$ and $f_{71} \sim f_{73}$) have $\Delta \simeq 0.0385$ and $\Delta \simeq 0.5711$, respectively. We denoise these feature values by dividing the corresponding $\Delta$, converting a data type from float to integer. The transformed integer features are expected to be more descriptive and informative when crossing features.

\subsection{Encoding Categorical Features}
The dataset includes many categorical features with high cardinality. We use frequency encoding and CatBoost encoding to avoid the curse of dimensionality. Frequency encoding, a.k.a. count encoding, replaces a category value with its number of occurrences for a given categorical feature. This simple strategy can be effective when the frequency is somewhat related to the target variable. We compared the following three alternatives for frequency encoding: (1) the number of occurrences during the previous day, (2) the number of occurrences during the previous week, and (3) the number of occurrences up to one day ago. We selected the second one due to its highest performance. Next, we conduct CatBoost encoding, a target encoding variation that supports time-aware encoding and regularization to avoid over-fitting. Similar to the sliding window, the category of the target day was transformed after fitting the encoder based on the data up to the previous day. The operations were performed for the two target variables, click and install, respectively.

%Catboost Encoding
%Catboost Encoding은 Ordered Target Encoding Method으로 현재의 instance를 포함하지 않은채 target을 기반으로 인코딩하여, 미래정보를 안봄? / leaakge 강한 느낌 
%총 2개의 target인 click / install 두가지를 기반으로 수행
%1일 전까지의 데이터를 기반으로 인코딩을 학습하여, 다음날 인코딩하는 방법을 택함 
% sliding window 방식으로

% f_1으로 정렬후에 Catboost Encoding 수행 

%CatBoost encoding is an alternative for target encoding that may cause overfitting.
%Frequency Encoding (혹은 Count Encoding)은 카테고리를 몇번 등장했는지 빈도수를 기반으로 인코딩하는 방법으로 기존 카테고리 값이 Sparse 하거나 시간에 따라 많이 변할때 의미가 있다. 
%우리는 학습기간과 평가기간 동안의 빈도수가 일관성있게 아래의 3가지 형태의 빈도 수를 실험했고 그 중 가장 성능이 높았던 2번 방식을 택했다. 
%(1) 직전 1일동안의 카테고리 빈도 수 
%(2) 직전 1주일동안의 빈도 수 
%(3) 1일 전까지의 등장한 카테고리의 빈도 수
%특히, 매시점마다 1주일 기간의 instance의 수가 달라져서 빈도수가 일정하지 않는 문제를 방지하고자 빈도수로 인코딩 된 값에 1주일 기간의 Instance의 총 합으로 나눠서 스케일링을 취해주었다. 

%\subsection{CatBoost Encoding}

% feature selection + frequency encoding, catboost encoding

\section{Models}
% 자세한 hyperparameter setting detail은 github repo 참조하라고 해야 함. depth, num_rounds 등 기본적인 파라미터만 공유
% num_leaves : 491 / max_depth : -1 / learning_rate : 0.006883242363721497 / gbdt / num_rounds 10000 with early_stopping 100
We employ GBDT as our prediction model. Specifically, we use LightGBM, which has been proven to show state-of-the-art performance in many previous competitions \cite{recsys19, recsys22}. We also tested other baseline models known to perform well in predicting tabular data, such as TabNet \cite{tabnet} and CatBoost. However, LightGBM was more robust and performed the best in our experiments. The tuned hyperparameters of LightGBM are as follows: \textit{number of leaves} = $491$, \textit{max depth} = $-1$, \textit{boosting type} = gbdt, and \textit{number of iterations} = $10,000$ with \textit{early stopping rounds} = $100$. Note that we did not apply an ensemble technique that blends the prediction results of several different models because we focus on a lightweight solution for rapid prototyping.

%사용한 모델들 간단히 d
%Tuning hyperparams? optuna?
% optuna 튜닝한 것은 떨어져서 옛날에 쓰던 hyp 사용했습니다 

%Our source code is publicly available: \href{https://github.com/choco9966/recsys-challenge-2023}{https://github.com/choco9966/recsys-challenge-2023}.

\section{Experiments}
%tabnet : 리더보드 6.221323
%catboost : 6.106765
%lightgbm : 6.059065
%- LightGBM / Catboost / Tabnet 3개 사용 결과 작성
%- Ensemble 했는데 LB 성능이 오히려 떨어졌다.
% 테스트용도로 넣어봄

\begin{table}[t!]
\caption{Performance of the prediction models. The best-performance model is denoted in bold.}
\label{tbl:performance}
\centering
\begin{tabular}{lrr}
\toprule
Model & Local Validation Score & Leaderboard Score\\
\midrule
(1) Vanilla LightGBM & 0.375506 & 6.466970\\
(2) = (1) + Frequency Encoding & 0.368485 & 6.264767\\
(3) = (2) + Denoisng Features & 0.361400 & 6.218796\\
(4) = (3) + CatBoost Encoding + Additional Feature Engineering & 0.361309 & 6.071975\\
\textbf{(5) Final LightGBM = (4) + Hyperparameter Tuning} & \textbf{0.360991} & \textbf{6.059065}\\
(6) CatBoost & 0.359850 & 6.106765 \\
(7) TabNet & 0.369832 & 6.221323 \\
\bottomrule
\end{tabular}
\end{table}

\begin{figure}[t!]
\centering
\includegraphics[width=0.80\linewidth]{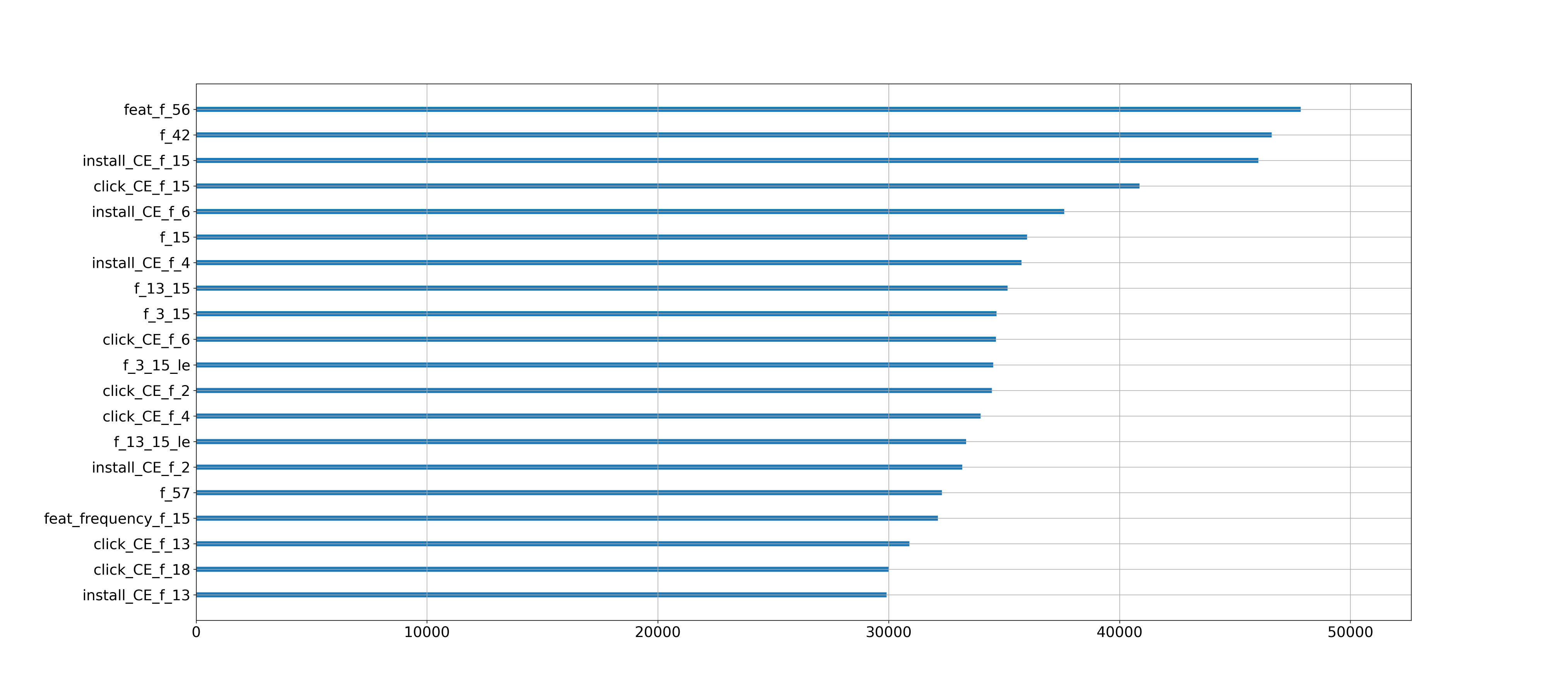}
\caption{LightGBM feature importance plot based on the number of tree splits.}
\label{fig:feat-importance}
\end{figure}

To evaluate the effectiveness of our method, we conduct detailed experiments and illustrate the results in Table \ref{tbl:performance}. By incorporating validated features, we can observe an improvement in the performance of our model. It is worth noting that CatBoost reported a good performance on local validation but poor performance on the leaderboard. Although both LightGBM and CatBoost belong to GBDT, in our study, LightGBM showed a much more robust performance than CatBoost. Fig. \ref{fig:feat-importance} shows the top 20 important features according to the LightGBM feature importance measure based on the number of times the feature is used in tree splitting. All experiments were conducted on virtual machines with 64 vCPUs, 120GB RAM, and NVIDIA RTX 3090 GPU.

\section{Conclusion}
This paper presents our approach to the ACM RecSys Challenge 2023. Our solution comprises adversarial validation, feature engineering, and a prediction model. In the adversarial validation phase, we detect and exclude potential features exhibiting the covariate shift. Next, in the feature engineering phase, we focused on denoising noisy continuous features and transforming categorical features into informative continuous values by performing frequency or CatBoost encoding. We employ LightGBM as our primary prediction model with tuned hyperparameters, and the final model showed excellent performance on the leaderboard even without applying an additional ensemble method. The proposed lightweight solution is simple yet effective for the following reasons: 1) adversarial validation can filter out non-informative features in the early stage, and 2) it does not apply the ensemble method that often requires much effort to find the optimal configuration. We believe our solution can be used for rapid prototyping for CTR/CVR prediction tasks.

\bibliographystyle{ACM-Reference-Format}
\bibliography{main}

\end{document}